\documentclass[5p]{elsarticle}
\usepackage{lineno,hyperref}
\usepackage[utf8]{inputenc}
\modulolinenumbers[5]

\journal{Computers and Electronics in Agriculture}

\usepackage{algorithm2e}
\usepackage{graphicx}
\usepackage{subcaption}
\usepackage{float}
\usepackage{microtype}
\begin{document}

\begin{frontmatter}

\title{Efficient identification, localization and quantification of grapevine inflorescences in unprepared field images using Fully Convolutional Networks}

\author[Bonn]{Robert Rudolph}
\author[Grapevine]{Katja Herzog}
\author[Grapevine]{Reinhard Töpfer}
%\author[Bonn]{Volker Steinhage}
\author[Bonn]{Volker Steinhage\corref{cor1}}
\ead{steinhage@cs.uni-bonn.de}
\cortext[cor1]{Corresponding author, Tel.: +49 (0)228 734538}


\address[Bonn]{Department of Computer Science IV, University of Bonn, Endenicher Allee 19A, D-53115 Bonn, Germany}
\address[Grapevine]{Institute for Grapevine Breeding Geilweilerhof, Julius Kühn-Institut, Federal Research Centre for Cultivated Plants, Siebeldingen, Germany}

\begin{abstract}
Yield and its prediction is one of the most important tasks in grapevine breeding purposes and vineyard management.
Commonly, this trait is estimated manually right before harvest by extrapolation, which mostly is labor-intensive, destructive and inaccurate.
In the present study an automated image-based workflow was developed quantifying inflorescences and single flowers in unprepared field images of grapevines, i.e. no artificial background or light was applied.
It is a novel approach for non-invasive, inexpensive and objective phenotyping with high-throughput.

First, image regions depicting inflorescences were identified and localized.
This was done by segmenting the images into the classes ‘inflorescence’ and ‘non-\-in\-flo\-res\-cence’ using a Fully Convolutional Network (FCN).
Efficient image segmentation hereby is the most challenging step regarding the small geometry and dense distribution of flowers (several hundred flowers per inflorescence), similar color of all plant organs in the fore- and background as well as the circumstance that only approximately $5\%$ of an image show inflorescences.
The trained FCN achieved a mean Intersection Over Union (IOU) of $87.6\%$ on the test data set.
Finally, individual flowers were extracted from the ‘inflorescence’-areas using Circular Hough Transform.
The flower extraction achieved a recall of $80.3\%$ and a precision of $70.7\%$ using the segmentation derived by the trained FCN model.

Summarized, the presented approach is a promising strategy in order to predict yield potential automatically in the earliest stage of grapevine development which is applicable for objective monitoring and evaluations of breeding material, genetic repositories or commercial vineyards.
\end{abstract}


\end{frontmatter}

%\linenumbers

\section{Introduction}

Grape yield is one of the most important traits in the scope of grapevine breeding, breeding research and vineyard management (Molitor et al. 2012 \cite{molitor2012impact}, Preszler et al. 2013 \cite{preszler2013cluster}, Töpfer \& Eibach 2016 \cite{topfer2016pests}, Simonneau et al. 2017 \cite{simonneau2017adapting}).
It is affected by genetic constitution of cultivars, training system, climatic conditions, soil and biotic stress (Bramley et al. 2011 \cite{bramley2011variation}, Kraus et al. 2018 \cite{kraus2018effects}, Howell 2001 \cite{howell2001sustainable}).
Several prediction models are recently published often based on destructive, laborious measurements and extrapolations right before harvest (detailed overview is given by de la Fuente et al. 2015 \cite{de2015comparison}).
For targeted vineyard management, i.e. yield adjustments due to bunch thinning, early yield predictions between fruit set and veraison (begin of grape ripening), are required in order to achieve well-balanced leaf-area-to-fruit-ratios, which are essential for maximized grape quality (Auzmendi \& Holzapfel 2014 \cite{auzmendi2014leaf}, de la Fuente et al. 2015 \cite{de2015comparison}).

Flower development, flowering and fruit set rate are directly linked to the amount of yield and thus are promising traits for comparative studies (Petrie et al. 2005 \cite{petrie2005effects}).
In grapevine breeding programs and research, investigations regarding the flower abscission (i.e. level of coulure or fruit set rate) and its genetic, physiological and environmental reasons are of peculiar interest (Boss et al. 2003 \cite{boss2003new}, Lebon et al. 2004 \cite{lebon2004flower}, Marguerit et al. 2009 \cite{marguerit2009genetic},  Giacomelli et al. 2013 \cite{giacomelli2013gibberellin}, Domingos et al. 2015 \cite{domingos2015flower}).
However, phenotyping of such small and finely structured traits is commonly done by visual estimations and thus achieve phenotyping scores that are more or less inaccurate and subjective, depending on the experience, awareness and condition of the employees.
Currently, more accurate measurements require much more labor-intensive and partially destructive measurements, which inhibits repetitive monitoring studies of several hundreds of different grapevine genotypes, e.g. crossing progenies (Giacomelli et al. 2013 \cite{giacomelli2013gibberellin}, Keller et al. 2010 \cite{keller2010spring}).

\begin{figure}
  \centering
    \begin{subfigure}[b]{0.97 \linewidth}
        \includegraphics[width=\textwidth]{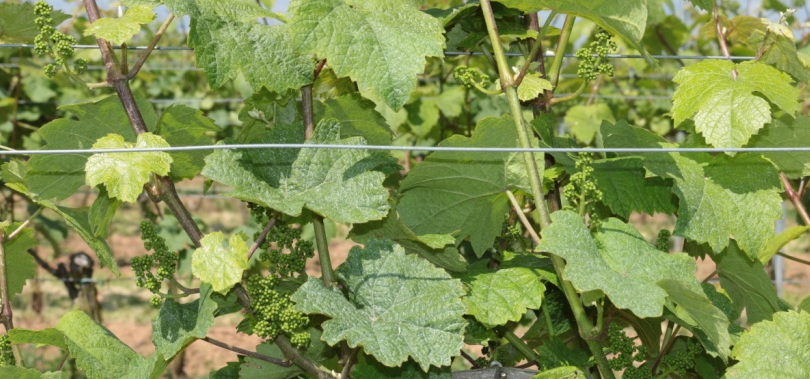}
        \caption{}
    \end{subfigure}
    \begin{subfigure}[b]{0.349\linewidth}
        \includegraphics[width=\linewidth]{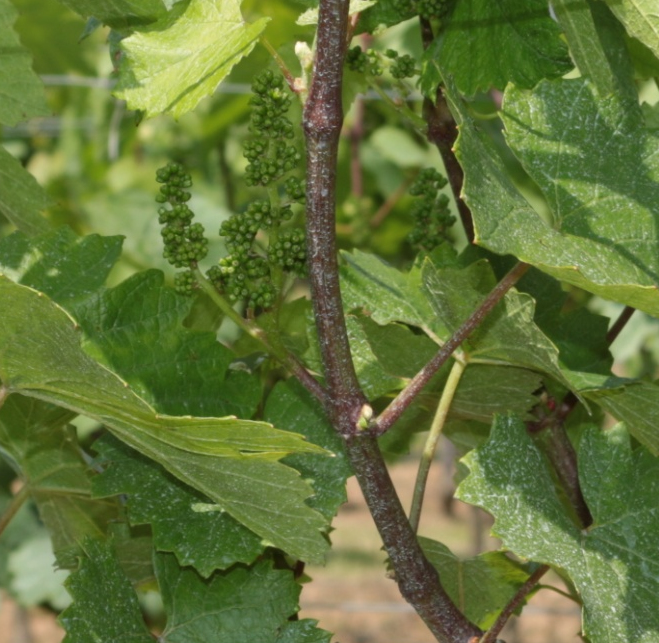}
        \caption{}
    \end{subfigure}
    \begin{subfigure}[b]{0.268\linewidth}
        \includegraphics[width=\linewidth]{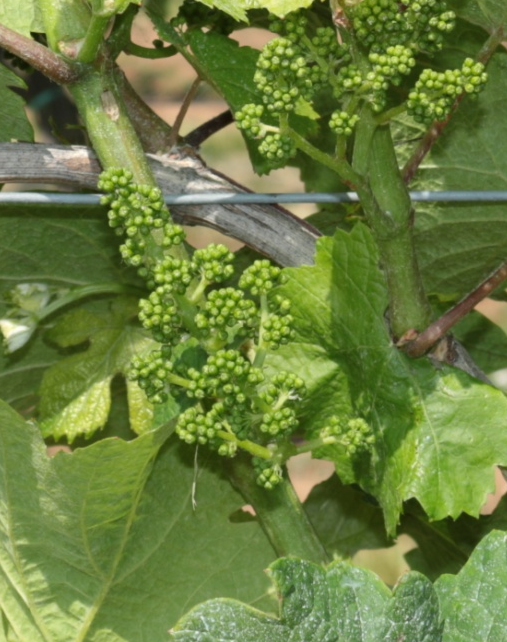}
        \caption{}
    \end{subfigure}
    \begin{subfigure}[b]{0.33\linewidth}
        \includegraphics[width=\linewidth]{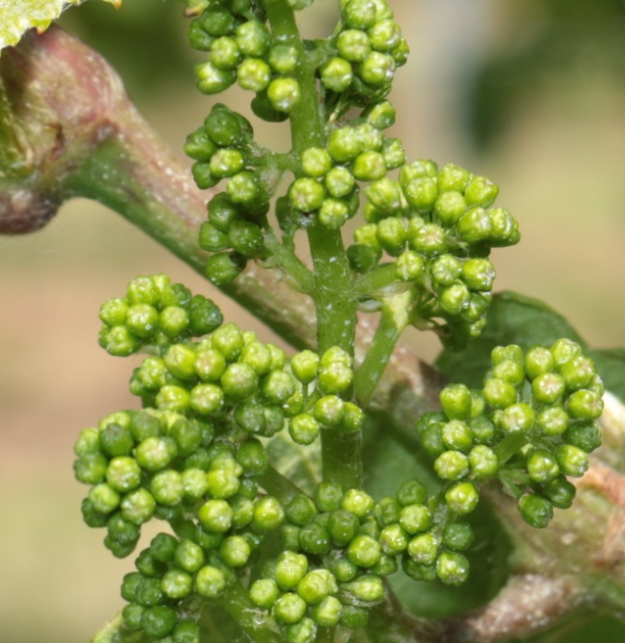}
        \caption{}
    \end{subfigure}
  \caption{Major challenges of unprepared grapevine images: Only 5\% of images are inflorescences and all plant organs are green (a); no standardized light conditions resulting in varying color characteristics of inflorescences, i.e. different green tones (b) (c); dense location of single flowers within one inflorescence (d).}
  \label{fig:introduction}
\end{figure}

The application of fast imaging sensors facilitates multiple field screenings of large experimental plots, breeding populations and genetic repositories.
In combination with efficient, automated data analysis objective, precise and comparable phenotypic data can be produced with minimal user interaction.
Fast, inexpensive and simple-to-apply sensors, e.g. consumer cameras, are promising for cost-benefit and user friendly approaches.
Recently, different sensor-based methods were developed for flower quantification based on images of individual captured grapevine inflorescences (Diago et al. 2014 \cite{diago2014assessment}, Aquino et al. 2015 a,b \cite{aquino2015vitisflower} \cite{aquino2015grapevine}, Millan et al. 2017 \cite{millan2017image}).
All of these approaches require images of a single inflorescence in front of well distinguishable backgrounds, i.e. artificial background or soil, which makes screenings of large plant sites more difficult and laborious.
Further, exertions of tractor-based approaches (Nuske et al. 2014 \cite{nuske2014automated}) or other field phenotyping platforms (Kicherer et al. 2015 \cite{kicherer2015automated}, Kicherer et al. 2017 \cite{kicherer2017phenoliner}, Aquino et al. 2018 \cite{aquino2018automated}) are not feasible by analysis of individual plant organs.
Regarding to this crucial restriction, a novel image analysis strategy for images of whole grapevine canopies without artificial background and additional light is required.
Efficient identifying and localizing of the image regions that depict inflorescences hereby is the most challenging step regarding the small geometry and dense distribution of flowers within inflorescences, the similar color of all plant organs in the fore- and background, as well as the circumstance that only approximately $5\%$ of the image area shows flowers (cf. fig. \ref{fig:introduction}).

In this study, the task of identifying and localizing the inflorescence areas was understood as a segmentation task, i.e. a task of partitioning the image into the classes 'inflorescence' and 'non-\-in\-flo\-res\-cence' by assigning a class label to each individual pixel.
While traditional approaches to image segmentation employ handcrafted heuristic criteria (e.g., intensity and color distributions) to identify appropriate image regions, deep learning convolutional neural networks (CNNs) allow learning descriptive criteria of the desired image regions just from the image data itself. 
By training a CNN on inflorescence segmentation data, a segmentation model able to deal with the complex scenes of images of whole grapevines was generated.
CNNs have established themselves as a state-of-the-art method for many tasks of image processing, including image classification (\cite{krizhevsky2012imagenet}, \cite{simonyan2014very}) as well as, more recently, image segmentation (\cite{long2015fully}, \cite{ronneberger2015u}).

CNNs used for image classification classify complete images, e.g. showing cars, buildings, dogs, etc. and generally follow a common structure that shows two phases: the feature extraction phase and the classification phase.
In the feature extraction phase multiple convolution layers and pooling layers generate successively more complex class characteristic image features (in the convolution layers) thereby downsampling the image size (in the pooling layers).
In the classification phase multiple fully connected layers derive class labels based on the derive image features.
CNNs for image segmentation generally implement a classification of each pixel in an image.
Two approaches to CNN-based image segmentations are most important here:

Long et al, 2015 \cite{long2015fully} introduced the Fully Convolutional Networks (FCNs) for image segmentation.
The architecture of a classification network is modified that way that its fully connected layers for the complete image classification are replaced by multiple convolutional layers and decoder layers.
In this network, the up-convolutional layers upsample the output size and the up-convolutional layers learn localization of class labels by combining the more precise high resolution features from layers of the extraction phase with the upsampled output.
Due to upsampling, this part can increase the spatial resolution up to the input-dimensions, providing per-pixel information on the input image.
Therefore, an FCN shows the following two phases: the feature extraction phase (as in the classification networks) followed by a decoder phase that results in a classification on the original image resolution, i.e. assigns a class label to each pixel of the image.
This kind of network is also called encoder-decoder network: a given input image is encoded in terms of features at different scales in the first phase while the second phase decodes all these features and generates a segmentation of the image.

U-Net (Ronneberger et al., 2015 \cite{ronneberger2015u}) is a popular architecture of FCNs and can be trained end-to-end.
This was not possible for the FCN approach presented by Long et al., 2015, which requires the encoder part to be pre-trained before being able to train the decoder part.

In this study, a U-net-like architecture of an FCN was used for the segmentation.

\begin{figure*}
  \centering
  \includegraphics[width=\linewidth]{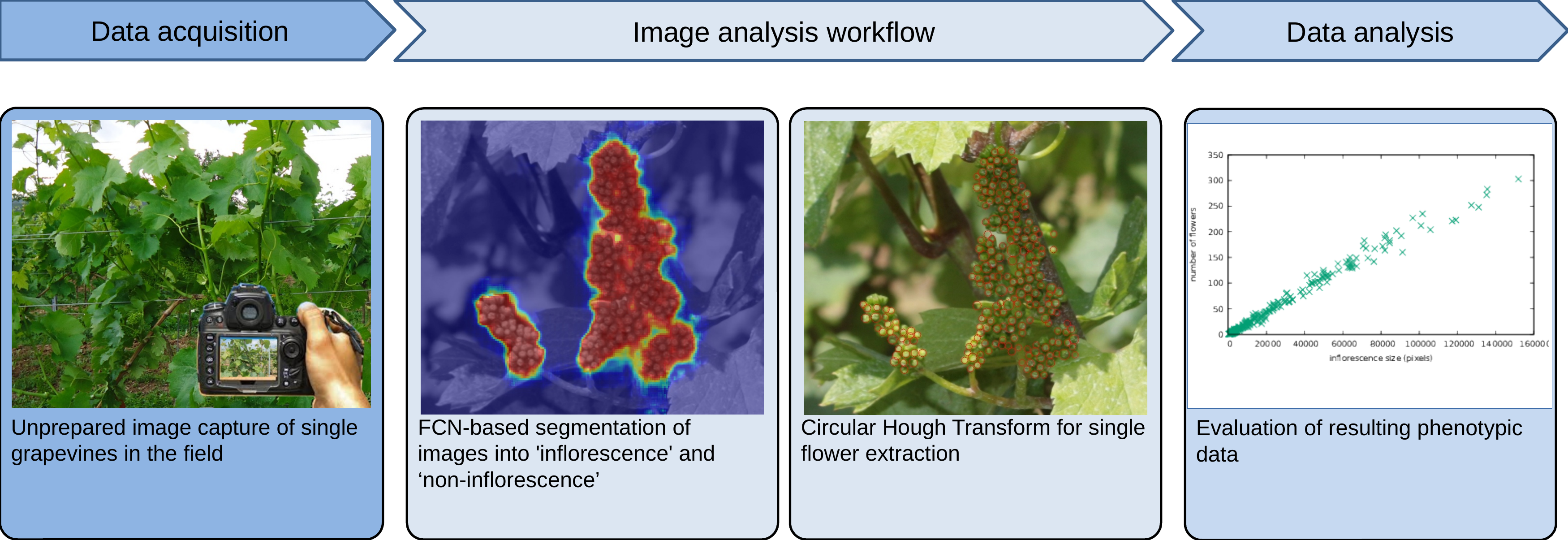}
  \caption{Phenotyping workflow: Captured images of grapevines are analyzed by segmenting them into 'inflorescence' and 'non-\-in\-flo\-res\-cence' via FCN and applying circle detection for flower extraction within the class 'inflorescence'. Finally, objective and precise phenotypic data are provided for further analysis.}
  \label{fig:workflow}
\end{figure*}

The major objective of the present study was the development and validation of an automated image analysis workflow in order to quantify the number of flowers per grapevine inflorescences in unprepared field images for non-invasive, inexpensive and objective phenotyping with high-throughput. The workflow of our approach shows four steps (see figure \ref{fig:workflow}):

\begin{samepage}
\begin{enumerate}
  \item Fast, inexpensive and simple-to-handle image taking with consumer camera.
  \item Identification and localization of inflorescences employing FCN-based image segmentation.
  \item Flower extraction by applying Circular Hough Transformation on segmented images.
  \item Evaluation of resulting phenotypic data
\end{enumerate}
\end{samepage}

\section{Materials and Methods}

\subsection{Plant Material, Image capture and pre-processing steps}
\label{sec:imagedata}

For image capturing, i.e., step 1 of our workflow (cf. fig. \ref{fig:workflow}), a single-lens reflex (SLR) camera (Canon EOS 70D) and a focal length of $35 mm$ was used for image capture in the field under natural illumination conditions with manually controlled exposure.
The distance to the plants was approximately $1 m$.

108 field images of the \textit{Vitis vinifera} ssp. \textit{vinifera} cultivars 'Riesling' and 'Chardonnay' were captured at the end of May 2016 when plants reached the BBCH stage 59 (Biologische Bundesanstalt, Bundessortenamt und CHemische Industrie; LORENZ et al. 1995 \cite{lorenz1995growth}).

\begin{figure}
\centering
\includegraphics[width=0.99\linewidth]{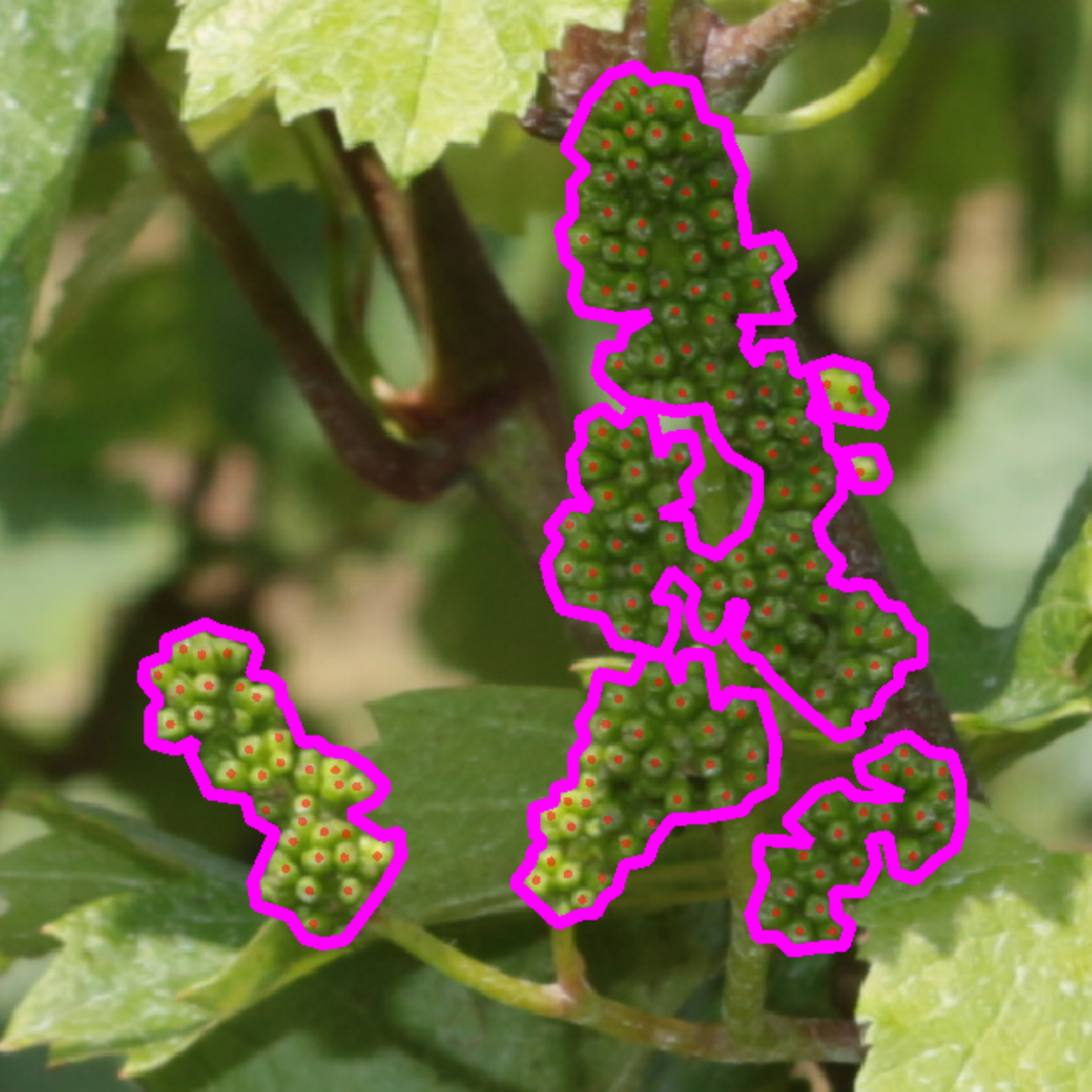}
\caption{Annotation of an inflorescence using bounding polygons (fuchsia) and of individual flowers using points (red) in the test set. The training set inflorescence polygons were annotated less precise in order to reduce workload.}
\label{fig:annotation}
\end{figure}

The field images were randomly divided into a training (98 images) and an evaluation (10 images) set.
Both training and evaluation set were manually annotated with bounding polygons around their inflorescences (see figure \ref{fig:annotation}).
Additionally, the evaluation set was annotated with the center of each individual flower visible within the images.

\subsection{Methodology}

Step 2 of our workflow (cf. fig. \ref{fig:workflow}) addresses the identification and localization of inflorescences in the image.

This is done by applying a trained FCN, as introduced in \cite{long2015fully} (cf. section \ref{sec:roisegmentation}), to the input image.
The FCN was trained on the annotated inflorescence segmentation data of the training set (cf. section \ref{sec:imagedata}).
After training, the FCN is able to derive a segmentation of the input image, determining for each pixel whether it is part of an inflorescence, or if it can be ignored.
Generally, the pixels depicting inflorescence form coherent regions.
Since these regions are of interest for further processing (i.e., flower extraction and deriving phenotyping data) these regions are called "regions of interest" (ROI).
In fig. \ref{fig:workflow} the result of an image segmentation is depicted in terms of a heat map where the red areas show the identified ROIs, i.e. the inflorescences.

In the third step of our workflow individual flowers are extracted from all detected ROIs of an image.
This was done by applying a Circular Hough Transform (CHT) on the image areas of the ROIs (cf. section \ref{sec:flowerextraction}).
For this study, the CHT was modified to consider the gradient direction, similar to the modification presented in \cite{roscher2014automated}.

\subsubsection{ROI segmentation}
\label{sec:roisegmentation}

Due to large areas of the image containing non-\-in\-flo\-res\-cence parts of vines and varying lighting conditions throughout the images, filtering the image by the color of individual pixels, as it was done by \cite{aquino2015grapevine} as a first step of finding inflorescences, would not produce reliable results.

\begin{figure}
\centering
\includegraphics[width=0.99\linewidth]{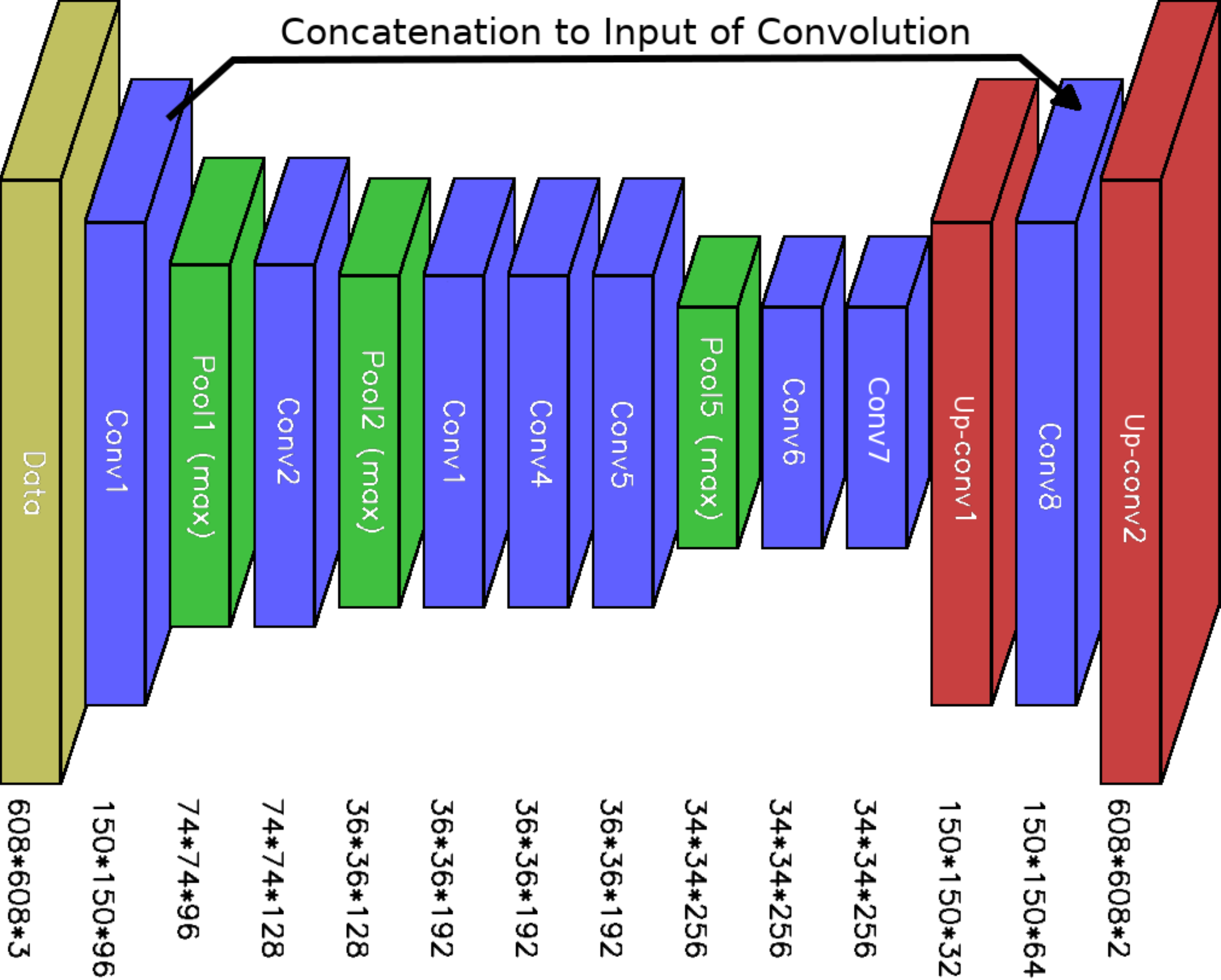}
\caption{The AlexNet-based FCN with up-convolution-based decoder part. Below spatial resolution and number of outputs are shown. The arrow denotes the concatenation of the channels of the outputs of $Conv1$ and $Up-conv1$, as input for $Conv8$. This is a visualization of the architecture presented in table \ref{tab:architecture}.}
\label{fig:architecture}
\end{figure}

Instead, we employ an U-net-like architecture, as proposed by Ronneberger et al, 2015 \cite{ronneberger2015u}, to identify and localize inflorescences in the image.
The network architecture is based upon the AlexNet architecture (\cite{krizhevsky2012imagenet}) as an encoder part, with a short decoder part added to it.
The architecture is visualized in fig. \ref{fig:architecture}, where each vertical blocks depicts layers of the full convolutional network.
The first olive block labeled as "$data$" depicts input images of size 608 x 608 with three layers for the RGB components.
The second, blue block labeled as "$Conv1$" depicts 96 convolutional filters.
Each filter encodes the image data in different types of trained features and shows an output size of 150 x 150 pixels.
The third, green block labeled as $Pool1$ (max) depicts the downsampling of the output of the 96 layers convolutional filters down to 74 x 74 pixels.
All following blue and green boxes up to the box labeled as "$Conv7$" depict convolutional and pooling layers that encode the input image successively in more and more complex and abstract feature representations, thereby downsampling the output size to 34 x 34 units.
This downsampling part can be seen as the left (downgoing) part of the shape of the character "U".
The following layers labeled as "Conv8", "Up-Conv1" and "Up-Conv2", respectively, show the upgoing part of the shape the character "U".
From this the name of the so-called U-net architecture was given by Ronneberger et al. \cite{ronneberger2015u}.
The upgoing part fuses and upsamples all feature representations to the original image size, thereby deriving class labels (inflorescence / non-\-in\-flo\-res\-cence) for all pixels.
For the sake of completeness, table \ref{tab:architecture} provides more detailed information for those who are familiar with convolutional networks and interested in the technical design of our U-net-based architecture.

For this network, the number of outputs of most layers were reduced from the values used in the AlexNet architecture in order to reduce memory requirements.

For the implementation the caffe-segnet (\cite{badrinarayanan2015segnet}) fork of the caffe library (\cite{jia2014caffe}) was used.

\begin{table}[hbt]
\begin{tabular}{llllll}
  \hline 
  \textbf{name} & \textbf{type} & \textbf{output}           & \textbf{ksize}    \\
  \hline
  Data          & Input         & $3 \times 608 \times 608$   &                 \\
  Conv1         & Convolution   & $96 \times 150 \times 150$  & $11 \times 11$  \\
  Pool1         & Max. Pooling  & $96 \times 75 \times 75$    & $3 \times 3$    \\
  Conv2         & Convolution   & $128 \times 75 \times 75$   & $5 \times 5$    \\
  Pool2         & Max. Pooling  & $128 \times 37 \times 37$   & $3 \times 3$    \\
  Conv3         & Convolution   & $192 \times 37 \times 37$   & $3 \times 3$    \\
  Conv4         & Convolution   & $192 \times 37 \times 37$   & $3 \times 3$    \\
  Conv5         & Convolution   & $192 \times 37 \times 37$   & $3 \times 3$    \\
  Pool5         & Max. Pooling  & $192 \times 35 \times 35$   & $3 \times 3$    \\
  Conv6         & Convolution   & $256 \times 35 \times 35$   & $3 \times 3$    \\
  Conv7         & Convolution   & $256 \times 35 \times 35$   & $3 \times 3$    \\
  Up-conv1       & Up-convolution & $32 \times 150 \times 150$  & $14 \times 14$  \\
  Concat        & Concatenate   & $128 \times 150 \times 150$ &                 \\
  Conv8         & Convolution   & $64 \times 150 \times 150$  & $3 \times 3$    \\
  Up-conv2       & Up-convolution & $2 \times 608 \times 608$   & $12 \times 12$  \\
  Prob          & Softmax       & $2 \times 608 \times 608$   &                 \\
  \hline
\end{tabular}
\caption{The network structure of the FCN used for inflorescence segmentation by layers. For the last output layer the softmax function ($Prob$ layer) is used to map the resulting two output values for each pixel to probabilities for the two output classes (inflorescence and non-\-in\-flo\-res\-cence respectively). The $Concat$ layer combines the channels of the outputs of $Conv1$ and $Up-conv1$.}

\tiny{Technical note: all convolution layers are followed by ReLUs. As in the AlexNet definiton, Local Response Normalizations follow the $Pool1$ and $Pool2$ layers. Both up-convolutions use a stride of $4$, $Conv8$ uses a padding of $1$. All further parameters were chosen according to the Alexnet definition.}
\label{tab:architecture}
\end{table}

This inflorescence segmentation network uses only two upsampling layers (denoted by \cite{ronneberger2015u} as "up-convolution").
Layer $Up-conv1$ upsamples to the resolution of layer $Conv1$.
The outputs of layer $Up-conv1$ and layer $Conv1$ are then appended and a convolution is applied in layer $Conv8$.
Layer $Up-conv2$ then upsamples to the resolution of the input image and produces output of the two classes, inflorescence and non-\-in\-flo\-res\-cence.
This architecture design was chosen on the assumption, that the information of the first convolution is sufficient to find a fine separation between inflorescence and non-\-in\-flo\-res\-cence, given the context of the surrounding image was provided by the last layer of the encoder network.

The training was done on the per-pixel class information provided by the manual annotation.
Due to the high memory requirements, the network could not be trained on the full $5472 \times 3648$ pixel images.
Instead, the network was trained on $5292$ non-overlapping images patches of $608 \times 608$ pixels (as depicted in the olive input data layer of fig. 4) produced from the training set.
Section \ref{sec:fullsegmentation} describes how we resolved this problem of memory footprint.
The technical training parameters were chosen as follows:

\begin{tabular}{ r l }
  solver: & Stochastic Gradient Descent \\
  learning rate: & $5 \cdot 10^{-5}$ (fixed) \\
  momentum: & 0.9 \\
  weight decay: & $10^{-4}$ \\
  batch size: & 1 \\
  iteration size: & 1
\end{tabular}

Since the detection and localization of inflorescences results in regions of interest (ROI), mean Intersection Over Union (IOU) was used as a quality measure.
IOU is defined per class as the cardinality of the intersection of the detected areas and actual areas of a class divided by cardinality of the union of these areas.

\begin{equation}
  IOU(c) = \frac{|T_c \cap P_c|}{|T_c \cup P_c|}
\end{equation}

\begin{figure}
\centering
\includegraphics[width=.99\linewidth]{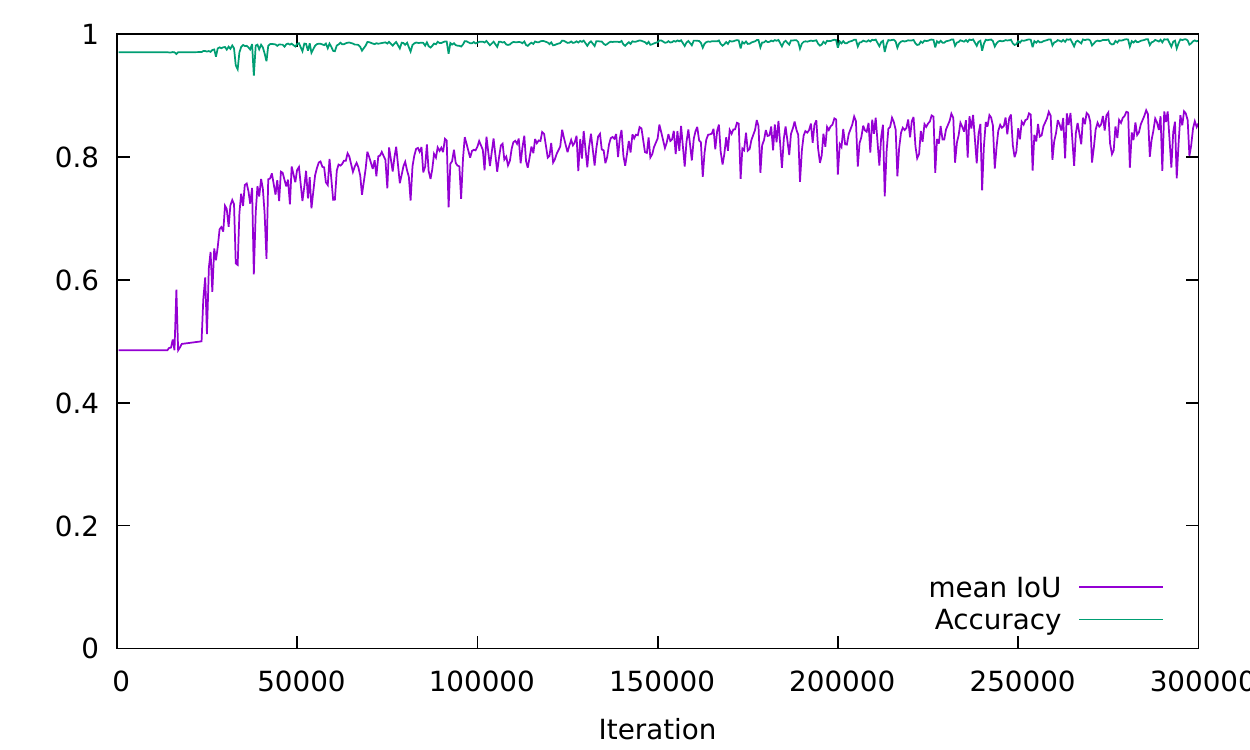}
\caption{Accuracy and mean Intersection Over Union (IOU) of the Fully Convolutional Network (FCN) during training. The flat behavior at the end of the graph indicates that further training would have not yielded much improvement of the model and that no overfitting occurred. The best-performing model was found after training for $285500$ iterations (IOU of $87.6\%$).}
\label{fig:training}
\end{figure}

The development of the mean IOU on the validation set during the training is shown in Figure \ref{fig:training}.
The best model produced by the training achieved a mean IOU of $87.6\%$ after $285500$ iterations.
This best-performing model was used as the segmentation model for the flower extraction.

\begin{figure*}
  \centering
    \begin{subfigure}[b]{0.9\linewidth}
        \centering
        \includegraphics[width=0.99 \linewidth]{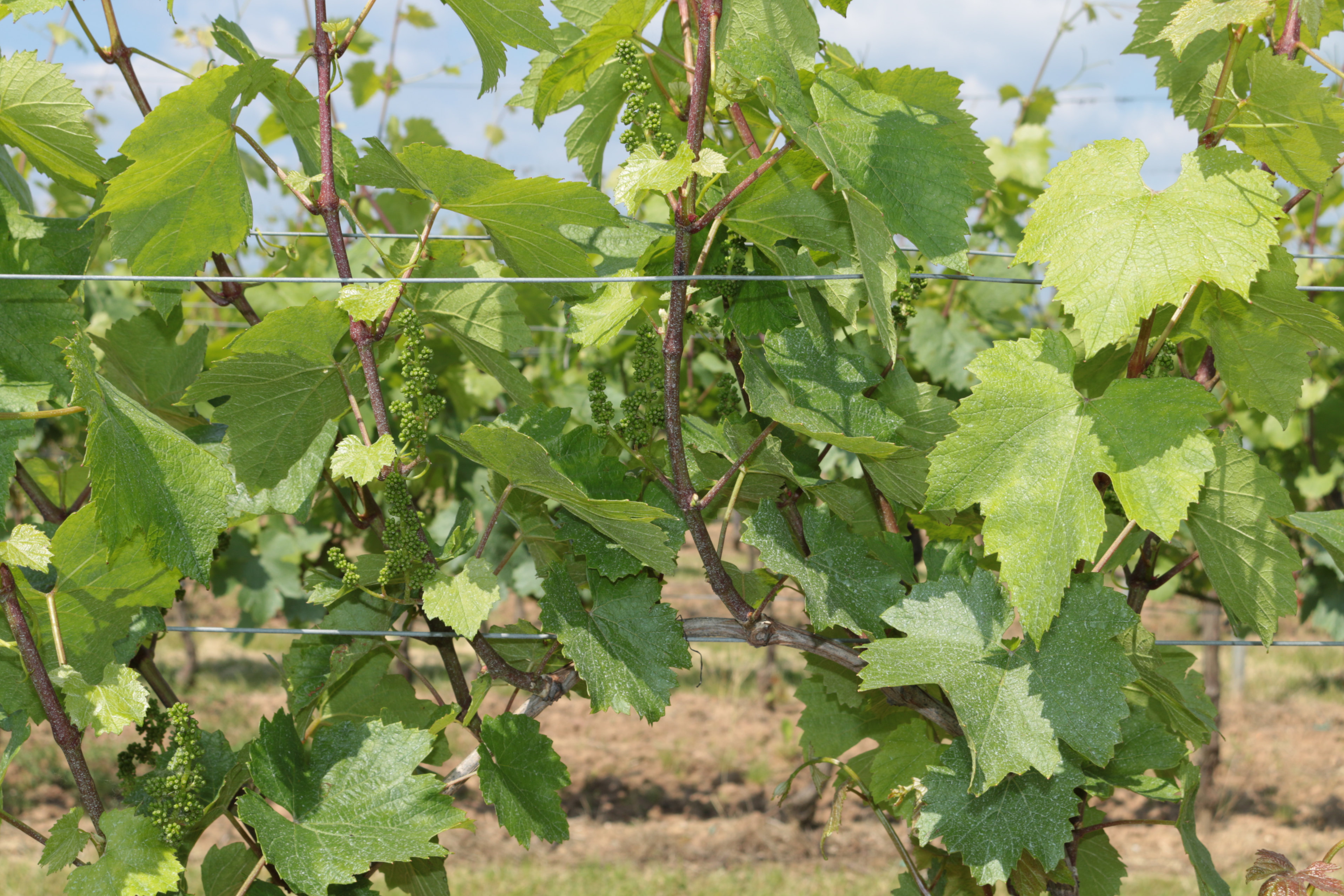}
        \caption{}
    \end{subfigure}
    \begin{subfigure}[b]{0.9\linewidth}
        \centering
        \includegraphics[width=0.99 \linewidth]{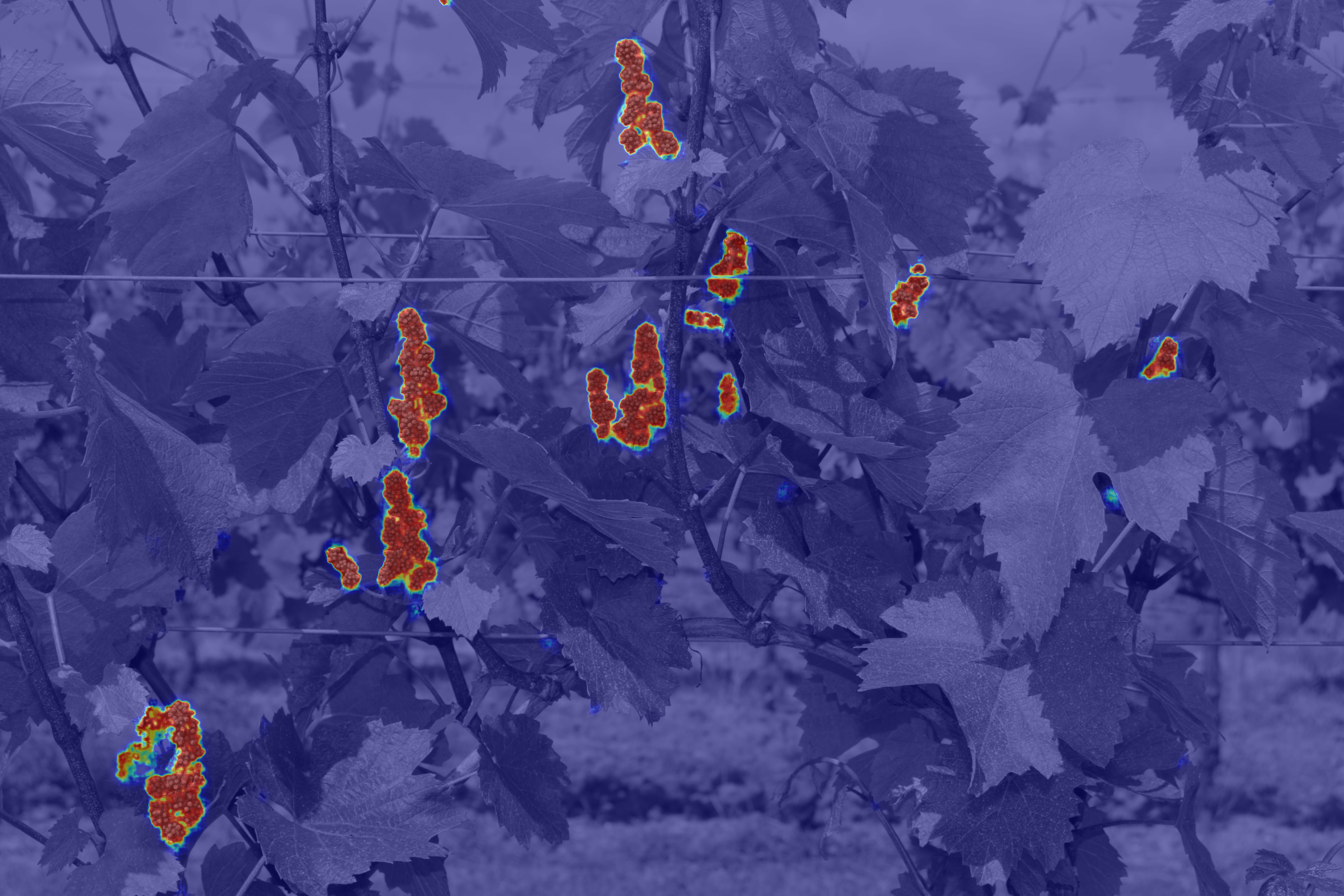}
        \caption{}
    \end{subfigure}
\captionsetup{width=0.9\linewidth}
\caption{Exemplary segmentation produced by the trained FCN model. (a) Original input image of the grapevine cv. 'Chardonnay'. (b) Segmentation heatmap with detected 'inflorescences'.}
\label{fig:segmentation}
\end{figure*}

Figure \ref{fig:segmentation} shows an example of the segmentations produced by the trained model.

\subsubsection{Full Image Segmentation}
\label{sec:fullsegmentation}

While the designed network architecture is fully convolutional and therefore can scale it's output with it's input, in practice this is not be possible with arbitrary high-resolution input sizes, due to memory requirements.
Instead of processing a complete image at once, it might be required to divide an image into smaller patches, as it was done previously for training the model.
For prediction, the segmentations produced for the patches then have to be recombined to produce a segmentation of the complete image.
This approach is a common workaround for this kind of bottleneck and was applied similarly in \cite{ronneberger2015u}.
The image patch size used during training does not limit the image patch size available for prediction.
If sufficient memory is available, a larger image patch size can be chosen in order to increase runtime performance, without requiring a new model to be trained.

\begin{figure}
  \centering
    \begin{subfigure}[b]{0.49\linewidth}
        \includegraphics[width=\linewidth]{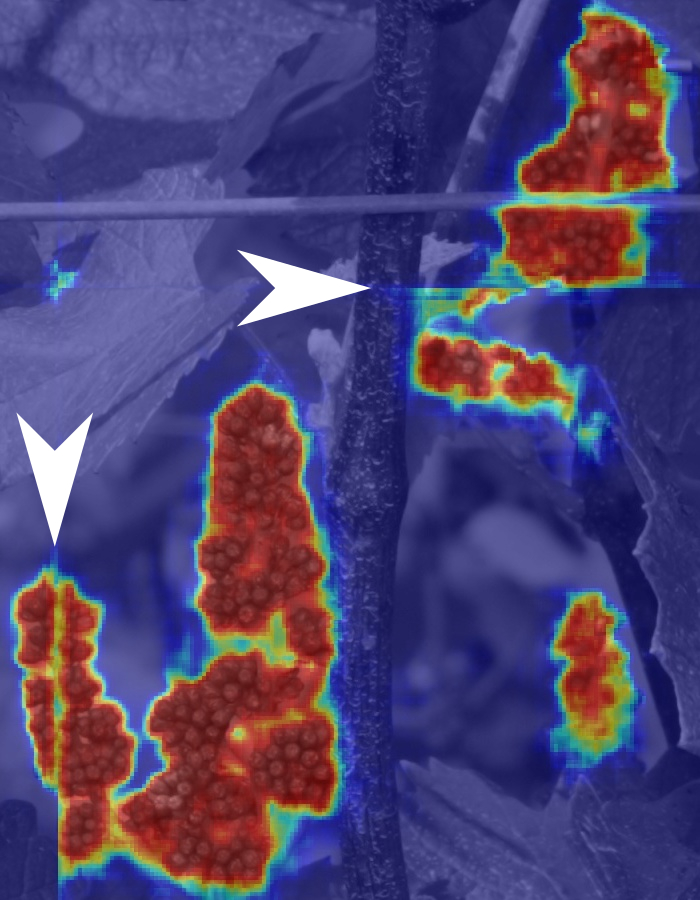}
        \caption{}
    \end{subfigure}
    \begin{subfigure}[b]{0.49\linewidth}
        \includegraphics[width=\linewidth]{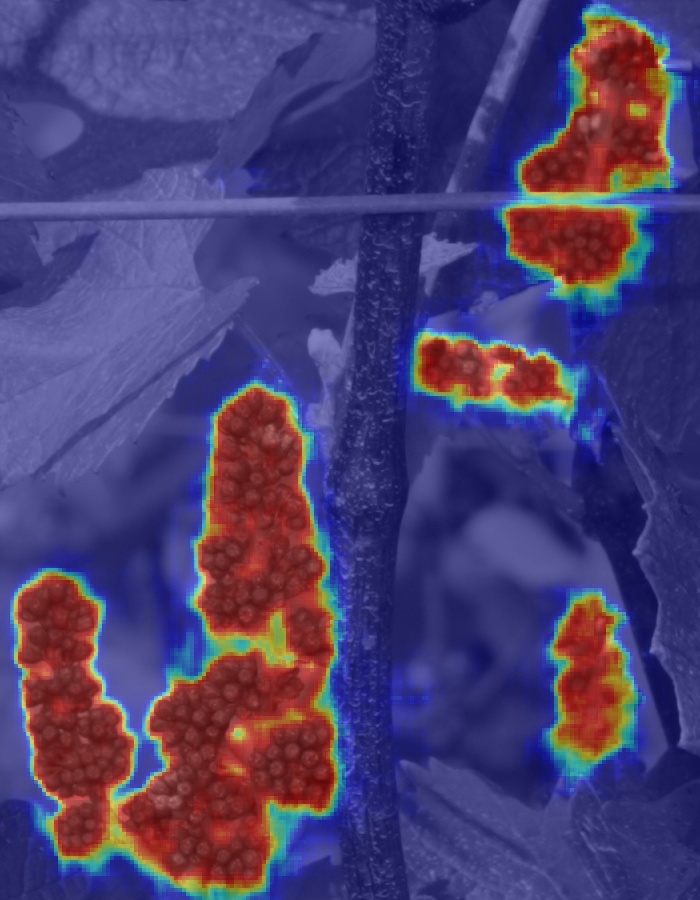}
        \caption{}
    \end{subfigure}
\caption{Assembly of FCN-segmented image patches into whole images. (a) Assembly of image patches without overlapping, resulting in artifact, inaccurate edges (white arrowheads). (b) Assembly of image patches with overlapping resulting in correctly segmented regions without missing information.}
\label{fig:overlapping}
\end{figure}

As opposed to \cite{ronneberger2015u}, the network used here was designed to produce an output of the same spatial size as the input by using padding and choosing the up-convolution kernels accordingly.
This resulted in the network producing sub-optimal results at the boundary edges between two patches.
This effect is shown in figure \ref{fig:overlapping}, in which segmentation errors can be seen along the edges between processed image patches.
As a workaround, the sub-optimal boundary edges of the prediction for each individual image patch were discarded.
This effectively results in the predicted segmentation covering a smaller area within the input image patch.
In order to produce a segmentation of the complete image the images were processed in overlapping patches in such a way that the smaller segmentations produced from the patches cover the complete input image.
At the boundary edges of the complete input image artificial context was provided by mirroring the input image.
This was done analog to the method described in \cite{ronneberger2015u}.

This results in a refined segmentation, which then can be used for the third step of our workflow, i.e. the flower extraction.

\subsubsection{Flower extraction}
\label{sec:flowerextraction}

For the extraction of flowers from the previously found ROIs we approximate the contours of the flowers by two-dimensional spheres.
Due to this approximation we can apply the Circular Hough Transform (CHT) to detect the flowers.
The CHT is a  well-established approach to find imperfect instances of spheres by a voting procedure that is carried out in the parameter space of two-dimensional spheres.
The parameter space of two-dimensional spheres shows three dimensions: two dimension for the two-dimension coordinates of the position (x,y) of center of a circle hypothesis and the third dimension for the radius of a circle hypothesis.

This approach is similar to that of \cite{roscher2014automated}, where CHT was used to find berries of grape vines and determine their size.
This third stage of processing of our workflow (cf. fig. \ref{fig:workflow} includes some pre-processing of the image, applying edge detection, removing any edges not within an ROI, applying the CHT and extracting the candidates according to the voting analysis in the parameter space.
The implementation was done using the OpenCV library (\cite{itseez2015opencv}).

For preprocessing a local contrast normalization, as described in \cite{jarrett2009best}, was applied, in order to allow for a single set of edge thresholds of the edge detection for flower contours throughout the image.
This was required, since different areas of the images could be more or less blurry, either due to depth of field and distance, or due to slight movement of the inflorescences by wind.

For edge detection the Canny Operator was used.
After using the implementation provided by the OpenCV library, edges not within an ROI were removed.

For the Hough Transform, flowers of the radii between domain-specific minimum and maximum values of flower radii were checked.
This interval represents the size of most flowers within the field images.
As a modification to the standard Hough Transform, each edge point only casts votes for a circle arc in its gradient direction, as well as the opposing direction, in order to reduce noise within the Hough Transform.
Here, the arc in which votes were cast was chosen at $\gamma = \frac{\pi}{8} = 22.5^{\circ}$.
Additionally the voting values were normalized by dividing them by the number of possible votes they could achieve in total, in order to allow for direct comparison between values of different radii.

For the extraction of candidate flowers all candidates above a certain threshold were sorted according to their value.
By iterating over this sorted list of candidates, starting with those of highest value, the final resulting flowers were selected.
This was done by maintaining an occupancy map.
If the center point of a candidate is not marked as occupied within the map, it is selected and a circle with $r = 1.5 \cdot r_{candidate}$ is marked as occupied on the map.
This increased radius was chosen to allow for slight overlapping of candidates.
After an iteration over all candidates, the selected candidates are returned as result.
This is shown in algorithm \ref{alg:selection}.

\begin{algorithm}
 \KwData{Sorted list of candidate circles C, Image size S, Radius factor a}
 \KwResult{A List of circles}
 Image $O$(S) := Unset\;
 $List Result \gets \emptyset$ \;
 \ForEach{$c \in C$}{%
  \If{$O(c.position) = Unset$}{%
   $Result$.append($c$)\;
   DRAW\_CIRCLE( $O$ , $c.position$,$c.radius \cdot a$)\;
  }
 }
 \caption{Algorithm for selecting the final circles ($Result$) from the circle candidates ($C$) produced by the Circular Hough Transform. By maintaining an occupancy map $O$ strongly overlapping circles are prevented.}
 \label{alg:selection}
\end{algorithm}

\begin{figure}[!ht]
\centering
    \begin{subfigure}[b]{0.421\linewidth}
        \includegraphics[width=\linewidth]{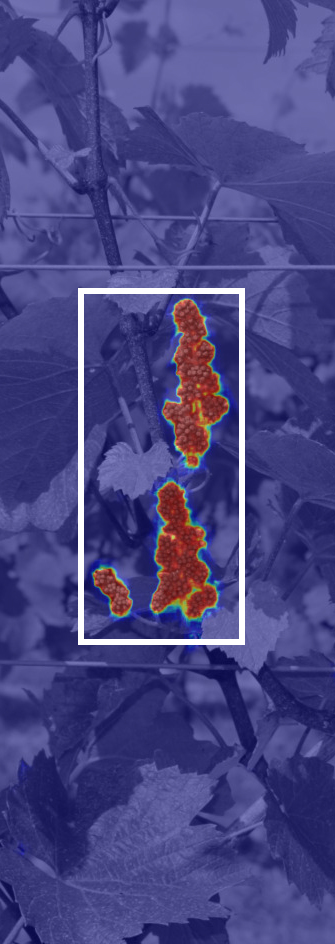}
        \caption{}
    \end{subfigure}
    \begin{subfigure}[b]{0.55\linewidth}
        \includegraphics[width=\linewidth]{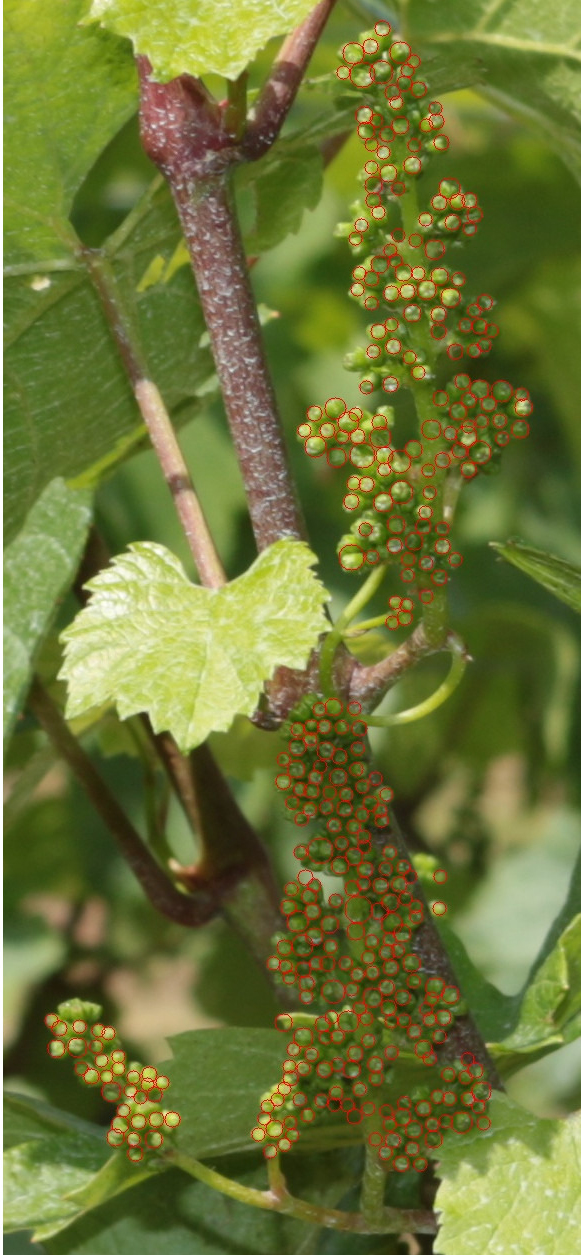}
        \caption{}
    \end{subfigure}
\caption{Sample of the Circular Hough Transform-based flower extraction, using the segmentation previously shown in figure \ref{fig:segmentation}. (a) Exemplary section of heatmap showing class 'Inflorescence', (b) result of flower extraction within the classified 'Inflorescence'.}
\label{fig:chtextraction}
\end{figure}

An example of this CHT-based flower extraction is shown in figure \ref{fig:chtextraction}.

For validation, the candidate flowers produced by the extraction were compared against the annotated flowers by iterating over the candidates and finding the closest annotated flower.
If a flower was within a certain radius-dependent distance of the candidate the candidate was considered a true positive.
Annotated flowers selected for one candidate were ignored for future candidates.
Candidates without a matching annotated flower were considered false positives and annotated flowers without a candidate  near them were considered false negatives.
Using these measures, F1 score, recall and precision were determined.

\section{Results}

The focus of this study is set on the efficiency of the image processing procedures, i.e., steps 2 and 3 of our workflow.
Therefore, we first present the evaluation of the identification and localization of inflorescenes by the trained FCN-segmentation model.
Then, we present the evaluation of flower detection and quantification.

\subsection{Identification and localization of inflorescenes}

\begin{figure}
  \centering
  \begin{subfigure}[b]{0.621 \linewidth}
    \begin{subfigure}[b]{\linewidth}
        \includegraphics[width=\linewidth]{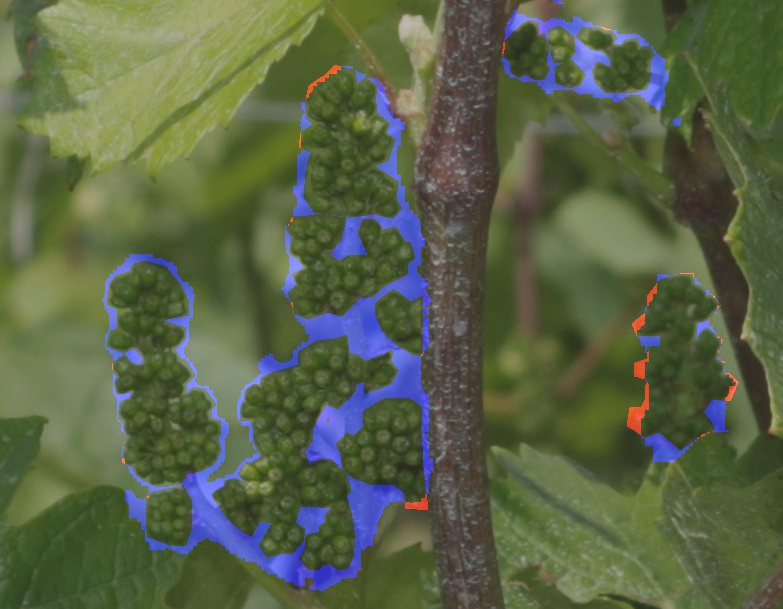}
        \caption{}
        \label{fig:errors_a}
    \end{subfigure}
    \begin{subfigure}[b]{\linewidth}
        \includegraphics[width=\linewidth]{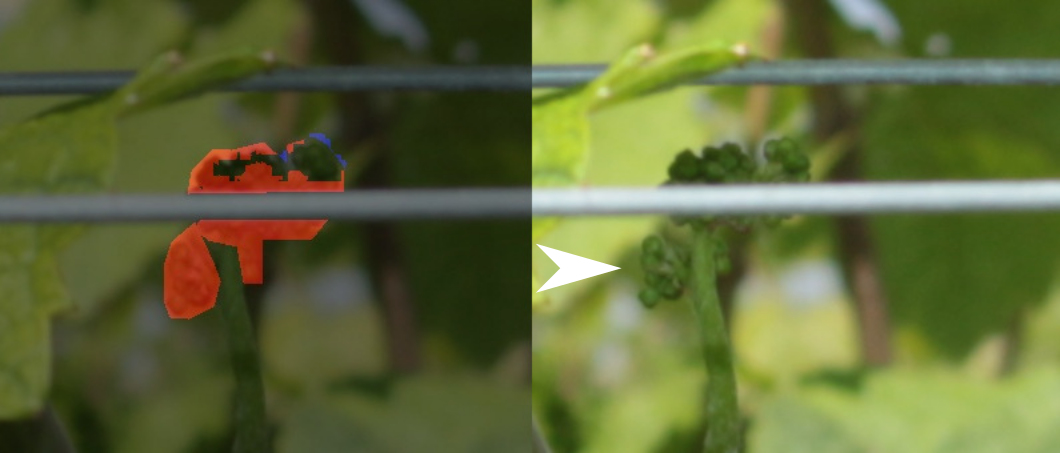}
        \caption{}
        \label{fig:errors_b}
    \end{subfigure}
  \end{subfigure}
  \begin{subfigure}[b]{0.279 \linewidth}
    \begin{subfigure}[b]{\linewidth}
        \includegraphics[width=\linewidth]{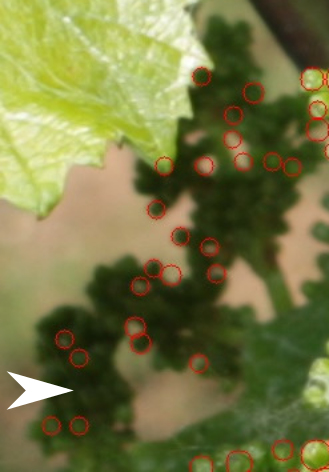}
        \caption{}
        \label{fig:errors_c}
    \end{subfigure}
    \begin{subfigure}[b]{\linewidth}
        \includegraphics[width=\linewidth]{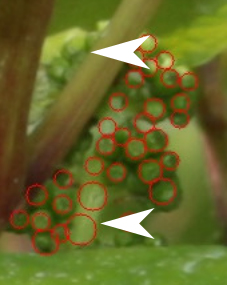}
        \caption{}
        \label{fig:errors_d}
    \end{subfigure}
  \end{subfigure}
  \caption{(a) False positives (blue) within an inflorescence and false negatives (red) at the edge of an inflorescence produced by the segmentation model. (b) False negatives (red) on a small inflorescence produced by the segmentation model. (c) Inconsistent flower extraction on a blurry background inflorescence detected by the segmentation. (d) False positives on a branch (lower arrowhead) and false negatives (upper arrowhead) of the flower extraction.}
  \label{fig:errors}
\end{figure}

The trained FCN-segmentation model achieved a mean Intersection Over Union (IOU) of $87.6\%$, with class-specific IOUs of $76.0\%$ for inflorescences and $99.1\%$ for non-\-in\-flo\-res\-cence.
Examining the segmentations on the test set predicted by the segmentation model, it can be shown that most wrong classifications are false positives occurring around actual flower areas (cf. fig. \ref{fig:errors_a}), while individual false positives occur rarely.
The false positives around flower areas are likely to be a result of the evaluation set being more precisely annotated than the training set.
False negatives occur more rarely and usually occur at very small inflorescences of only a few flowers (cf. fig. \ref{fig:errors_b}) and occasionally at the edges of larger inflorescences (cf. fig. \ref{fig:errors_a}).

While the FCN-segmentation model achieves high IOU values on the test set and in practice performs well as a basis for the flower extraction (cf. section \ref{sec:flowerextractionresults}), the performance could still be improved.
Training the model on more images would, most likely, allow the network to learn a better generalization.
Additionally, data augmentation (e.g. modified brightness, hue, saturation) during training could further improve the results.

Since the model was not yet tested on other grapevine varieties, it is unknown how well it generalizes to those.
However, it can be assumed that it performs best on the varieties it was trained on.
Incorporating more different grapevine varieties into the training set would improve the generalization of the inflorescence detection, allowing for the application on other varieties.

\subsection{Flower detection and quantification}
\label{sec:flowerextractionresults}

The flower detection and quantification was evaluated separately on (a) the manually generated ground truth segmentation, (b) the segmentation produced by the trained model and (c) on the complete image without providing segmentation.
The performance measures F1 score, Recall and Precision using each of the segmentations are shown in table \ref{tab:extraction-results}.
Additionally, the 'EOA' column shows the mean amount of flowers estimated over the amount annotated.
The standard deviation of this value over the validation set is given in the '$\sigma$(EOA)' column.

\begin{table*}
\begin{tabular*}{\textwidth}{llllll}
  \hline 
  \textbf{Segmentation}       & \textbf{F1}  & \textbf{Recall} & \textbf{Precision} & \textbf{EOA} & \textbf{$\sigma$(EOA)} \\
  \hline
  None                        &   9.8\%      &   85.5\%        &   5.2\%            &   1718.7\%     & 336.7\% \\
  Segmentation Model          &   75.2\%     &   80.3\%        &   70.7\%           &   115.4\%      & 8.66\% \\
  Ground Truth                &   80,0\%     &   84.3\%        &   76.1\%           &   112.7\%      & 10.17\% \\
  \hline
\end{tabular*}
\caption{The performance of the flower extraction using different segmentations. The performance measures include F1 score, Recall, Precision, mean estimated over actual number of flowers (EOA) and standard deviation of EOA over the test set.}
\label{tab:extraction-results}
\end{table*}

The flower extraction is prone to producing false positives over false negatives, generally resulting in a higher recall than precision.
False negatives can occur in regions not labeled as ROI by the previous step (cf. fig. \ref{fig:errors_d}, top) and at inflorescences which were labeled as ROI, but which are too blurred for the CHT to detect the individual flowers (cf. fig. \ref{fig:errors_c}).
False positives often occur on other small plant structures within ROIs, e.g the stems of inflorescences (cf. fig. \ref{fig:errors_d}, bottom).

\begin{figure}
\centering
\includegraphics[width=.8\linewidth]{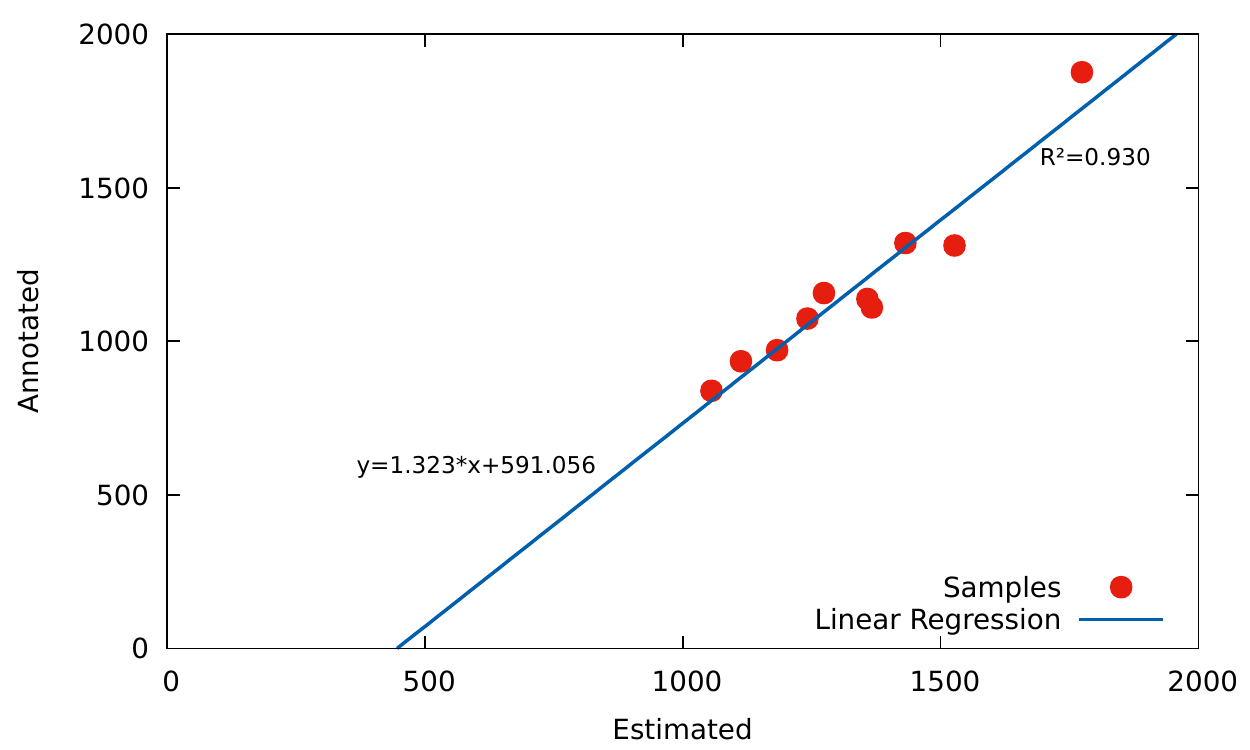}
\caption{Linear regression model of estimated vs. annotated flower numbers within the test set. The test set consists of 10 randomly chosen images (6 Chardonnay, 4 Riesling) of differing perspectives.}
\label{fig:linear-regression}
\end{figure}

Since the flower extraction tends to overestimate the number of flowers (see table \ref{tab:extraction-results}, column EOA), a linear regression was fitted to correct for it.
Table \ref{tab:testdata} shows these measurements as well as the absolute number of flowers estimated and annotated for each individual test sample for the segmentation model.
This linear model (shown in figure \ref{fig:linear-regression}) achieved a coefficient of determination of $R^2 = 0.930$.

\begin{table*}
\begin{tabular*}{\textwidth}{llllllll}
  \hline 
  \textbf{Image}       & \textbf{F1}  & \textbf{Recall} & \textbf{Precision} & \textbf{EOA} & \textbf{Annotated} & \textbf{Estimated} \\
  \hline
  Chard. Frontal 04           & 78.3\%       & 82.2\%          & 74.7\%             & 110.1\%      & 1157        & 1274 \\
  Chard. Frontal 08           & 76.4\%       & 86.2\%          & 68.5\%             & 125.8\%      & 839         & 1056 \\
  Chard. Frontal 11           & 74.3\%       & 80.1\%          & 69.3\%             & 115.6\%      & 1074        & 1242 \\
  Chard. Upwards 02           & 77.1\%       & 83.4\%          & 71.7\%             & 116.3\%      & 1312        & 1527 \\
  Chard. Upwards 05           & 72.1\%       & 70.2\%          & 74.2\%             & 094.5\%      & 1876        & 1774 \\
  Chard. Downwards 07         & 76.0\%       & 83.3\%          & 69.9\%             & 119.0\%      & 935         & 1113 \\
  Riesl. Frontal 09           & 75.8\%       & 83.1\%          & 69.6\%             & 119.3\%      & 1138        & 1358 \\
  Riesl. Upwards 04           & 72.6\%       & 81.0\%          & 65.8\%             & 123.1\%      & 1110        & 1367 \\
  Riesl. Upwards 13           & 72.9\%       & 80.9\%          & 66.4\%             & 121.8\%      & 971         & 1183 \\
  Riesl. Upwards 04           & 77.6\%       & 80.9\%          & 74.5\%             & 108.4\%      & 1320        & 1432 \\
  \hline
\end{tabular*}
\caption{The Performance of the flower extraction using the trained segmentation model for all images of the test set. The measures include F1 score, Recall, Precision, estimated over actual number of flowers (EOA), as well as the raw numbers of annotated and estimated flowers.}
\label{tab:testdata}
\end{table*}

While the best performance was achieved on the ground truth segmentation, the F1 score on the segmentation produced by the trained model is lower by only $4.8\%$.
This relatively small difference in performance of the flower extraction using the automatically generated segmentation data and the annotated segmentation data shows that the application of an FCN-based segmentation model is a promising strategy.

To meet the challenge of overestimation in the flower extraction step, future work will investigate the employment of a trained CNN in the flower extraction step instead of the Circular Hough Transform.
The trade-off will be additional annotation work in the training data: instead of labeling inflorescences by bounding polygons (cf. fig. \ref{fig:annotation}) single flowers must be labelled - e.g. by bounding circles.
This new CNN-based flower extraction could be either implemented as a separate processing step operating on the inflorescence areas or directly as part of the segmentation network.

\subsection{Runtime Performance}

The runtime behavior of both the segmentation model and the flower extraction were evaluated on the following system: MSI GE60-OND Gaming Notebook; Intel Core i7-3630QM 2.4GHz, 8GB DDR3 RAM, GeForce GTX 660M (2GB GDDR5 SDRAM).

The operating system run was Arch Linux (64 bit Linux kernel, version \textit{4.12.10}). The libraries used were the at the time of writing most recent git version of \textit{caffe-segnet} (rc2-338-gdba43980), \textit{cuda 8.0.61}, \textit{cudnn 7.0.1} and \textit{opencv 3.3.0}.

The segmentation model was run using the GPU mode of the caffe library.
In order to make the best use of the massive parallelization possible with GPUs, the image patches were chosen as large as possible.
One of the spatially largest networks able to fit in the 2GB graphics memory of the test system was that of an input size of $1216 \times 1216$ pixels.
This allowed for processing of an complete image of $5472 \times 3648$ pixels in $20$ image patches.

Including mirroring at edges, disassembling into patches and reassembling, the mean time of a segmentation was measured at $7.8s$.
It can be expected that using upcoming, more modern graphics cards the runtime performance of the segmentation would significantly increase, due to availability of more memory, allowing for larger patches, as well as general increases in speed in modern hardware.

The flower extraction was run as a single-thread process on the CPU.
On the segmentation produced by the trained model the mean runtime per image was $4.161s$.
Since the memory footprint of the flower extraction is relatively low, when used in practice, the throughput of the flower extraction could be massively sped up by using more threads/cores.
This possibility makes the segmentation the main bottleneck of the complete system.

However, even without optimization of the flower extraction step, the total required time of about $12s$ per image should still allow for a practical application of the system.

\section{Conclusions}

In the present study, a low-cost and commercial available consumer camera was used in order to reveal simple-to-apply image acquisition of normal growth grapevines directly in vineyards.
Further, an efficient, automated image analysis was developed for reliable flower detection and quantification.
It is the first study facilitating efficient and contactless screening of large sets of grapevines receiving objective and high-quality phenotypic data.
This is important for further studies regarding the development of reliable early yield prediction models for objective characterization and multi-year monitoring of breeding material, e.g. crossing populations and genetic repository.
Early yield prediction is a promising strategy for grapevine training systems showing more complex canopy architectures, e.g. semi-minimal pruned hedges.
Further, the developed strategy makes carrying of artificial backgrounds or invasive treatments of grapevines due to defoliation redundant which opens up possible vehicle-based phenotyping applications.

Further, training the FCN on grapevine images of early stages of fruit development, i.e. fruit set (BBCH 71) or groat-sized berries (BBCH 73), will enable comparison of quantified flowers and quantified young berries in order to phenotype susceptibility to fruit abscission, i.e. level of coulure, objectively and with high throughput.
However, the system has to be robust for its reliable applications on high diversity phenotypes.
Therefore, further large data sets for different stages of plant development, different grapevine cultivars and phenotypic variable breeding populations are required for validation.

Finally, the development of an intuitive graphical user interface will improve usability for potential users, i.e. breeders or scientists.

\section*{Acknowledgements}

We gratefully acknowledge the German Research Foundation (Deutsche Forschungsgemeinschaft (DFG), Bonn, Germany (Automated Evaluation and Comparison of Grape\-vine Genotypes by means of Grape Cluster Architecture, STE 806/2-1, TO 152/6-1), as well as the German Federal Ministry of Education and Research (Bundesministerium für Bildung und Forschung (BMBF), Bonn, Gemany (NoViSys: FKZ 031A349E)).

\section*{References}

\bibliography{mybibfile}

\end{document}